\documentclass[letterpaper, 10 pt, conference]{ieeeconf}

\IEEEoverridecommandlockouts

\usepackage{graphicx}      
\usepackage{xcolor}        
\usepackage{booktabs}      
\usepackage{amsmath}       
\usepackage{amssymb}       
\usepackage{algorithm}    
\usepackage{algpseudocode} 
\usepackage{cite}          
\usepackage{xspace}        
\usepackage{balance}       
\usepackage{placeins}      
\usepackage[hidelinks]{hyperref}

\newcommand{\ours}{SonoGraph-WM\xspace}

\title{\LARGE \bf

Scanning While Imagining:\\

A Scene-Graph World Model for Robotic Ultrasound Navigation
}

\author{Xuesong Li$^{a,b}$, Shuai Chen$^a$, Feng Li$^{a,b}$, Zhongliang Jiang$^c$, Nassir Navab$^{a,b}$, and Yuan Bi$^{a,b}$\thanks{$^a$ Computer Aided Medical Procedures (CAMP), Technical University of Munich, Munich, Germany; $^b$ Munich Center for Machine Learning (MCML), Munich, Germany; $^c$ Department of Mechanical Engineering, The University of Hong Kong, Hong Kong SAR, China.}}

\begin{document}

\maketitle
\thispagestyle{empty}
\pagestyle{empty}

\bstctlcite{IEEEexample:BSTcontrol}

\begin{abstract}

Ultrasound (US) acquisition depends on the operator's ability to interpret
anatomy and anticipate how the view will change with probe motion.
Many robotic US navigation methods select actions without explicitly
predicting these anatomical changes. We propose \ours{}, an action-
and goal-conditioned world model for anticipatory probe navigation.
The model represents anatomy as scene graphs (SGs), capturing visible
structures, their geometry, and spatial relationships without synthesizing
US images. Given a history of SGs and probe poses, a unified Transformer
jointly predicts future SGs and poses. A receding-horizon planner
recursively imagines candidate trajectories, selects the shortest
predicted path reaching a goal graph, and follows it over a short
execution horizon before replanning from new observations.
To reduce reliance on tracked and anatomically annotated US sequences,
we generate aligned SG--pose training data from computed tomography
(CT) label maps along surface-constrained probe trajectories.
On four held-out CT cases, spatial relation F1 remains above 93\%
over 20 prediction steps, and closed-loop navigation achieves
77.50\% and 75.00\% success for the gallbladder and pancreas,
respectively, using annotation-derived SGs.
In robot--phantom navigation experiments with label-map-derived SGs,
the planner reached the target view in 73.7\% of trials.
These findings support CT-supervised anatomical world modeling for
probe planning and highlight the importance of frequent observation
updates for reliable navigation. Project Page: \url{https://noseefood.github.io/us-sonograph-wm/}.
\end{abstract}

\section{Introduction}

Medical ultrasound (US) is widely used for its portability and real-time imaging
without ionizing radiation~\cite{jiang2023robotic}. Obtaining the desired view requires
skilled adjustment of probe position and orientation, which can limit acquisition
reproducibility~\cite{jiang2023robotic}. Robotic US systems (RUSS) aim to improve
repeatability and reduce the physical workload of scanning~\cite{bi2024machine}.
During autonomous scanning, the robot must determine how to move the probe based
on the anatomy visible in the current US image.

\begin{figure}[t]
  \centering
  \includegraphics[width=0.85\columnwidth]{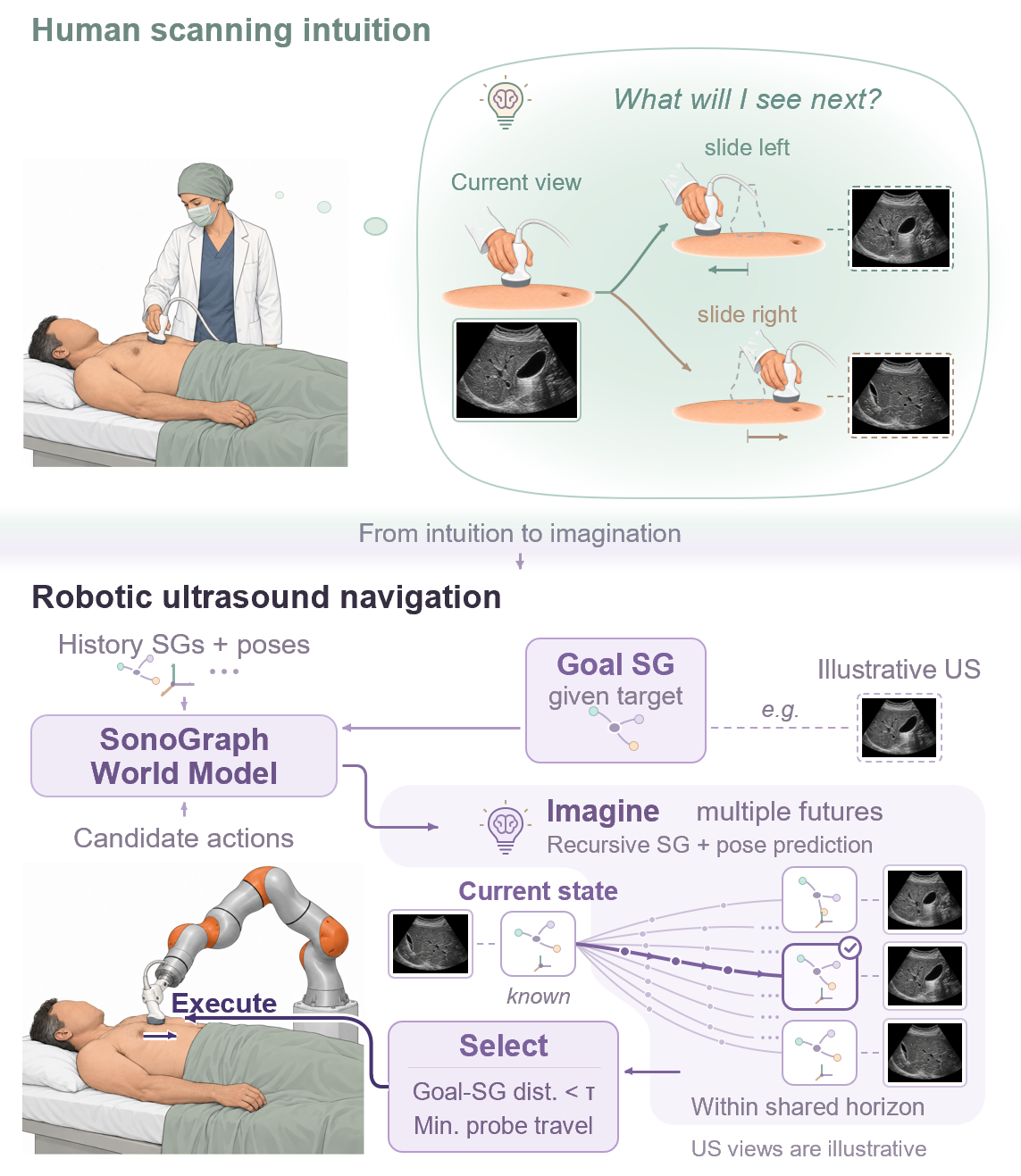}
  \caption{Concept overview of \ours. For each candidate probe motion, the model
           predicts anatomical scene graphs (SGs) and probe poses. The planner selects
           the shortest predicted path among those whose SGs match a goal graph.
           B-mode examples illustrate the anatomical views represented by the graphs.}
  \label{fig:concept}
\end{figure}

Existing RUSS navigation methods select probe motions in several ways.
Registration-based approaches use rib-cartilage geometry to align US with computed
tomography (CT) and transfer template scanning paths to individual
patients~\cite{jiang2023skeleton,jiang2025cartilage}. Rule-based and visual-servoing
methods~\cite{jiang2024needle} adjust the probe using detected anatomical
features~\cite{jiang2022tubular}. Model-free reinforcement learning (RL) and imitation
learning (IL) learn motions from interaction or expert
demonstrations~\cite{hase2020navigation,bi2022vesnet,jiang2025ultrabot}.
Rule-based controllers and model-free policies typically act without explicitly
predicting which structures will become visible. Such predictions would allow
candidate motions to be compared before execution, reflecting how sonographers
use anatomical knowledge to anticipate changes in the view.

Predicting these changes requires anatomical information beyond the current image.
B-mode US shows a cross-section rather than the complete three-dimensional
arrangement of organs. The target may lie outside the imaging plane, and probe
movement changes which structures are visible~\cite{li2025scenegraph}.
Sonographers use anatomical knowledge to interpret visible structures, guide
probe adjustments~\cite{jiang2023robotic}, and anticipate how neighboring
structures and their spatial arrangement may appear. We aim to give robots
this ability to anticipate anatomical changes before choosing their next motion.

A learned world model predicts what a robot would observe after an
action~\cite{ha2018world,hafner2019planet}. Recursive predictions enable comparison
of paths before execution, as explored in navigation and
manipulation~\cite{bar2025nwm,du2023unipi}. In US, DreamReg uses predicted observations
to refine registration~\cite{kang2026dreamreg}, while Fan et al.~\cite{fan2026acwm}
use rewards from predicted US images to train a predictor of probe motions
toward a target view.

However, predicting future US images requires modeling both anatomical changes
and appearance variations caused by speckle~\cite{li2025speckle2self}, low contrast,
and acquisition settings~\cite{li2025scenegraph}. Anatomical scene graphs (SGs)
describe structures without reproducing image texture.
Li et al.~\cite{li2025scenegraph} use SGs to explain US images and guide scanning
toward missing structures. Nodes encode anatomical structures, positions, and sizes,
while edges describe spatial relationships. This representation predicts which
structures candidate motions would reveal and where they would appear.
However, the guidance remains coarse and does not directly specify executable
robot motions. Extending it to motion planning requires learning how SGs evolve
with probe motion from aligned SG--pose sequences, which remain scarce in US.

Anatomically labeled CT volumes offer an alternative source of these sequences.
We sample simulated US imaging planes along generated probe paths using CADS
organ label maps~\cite{xu2025cads}. At each plane, intersecting structures, their
geometry, and spatial relationships form an SG, which is paired with the
corresponding probe pose. This produces multiple SG--pose sequences from
each volume without acquiring and annotating US scans for every path.
We use these sequences to learn anatomical changes under probe motion and
evaluate prediction and navigation using annotation-derived SG observations.

We train \ours{} on these sequences and use its predictions in a receding-horizon
planner (Fig.~\ref{fig:concept}). A unified Transformer takes the SG--pose
history, a candidate action for the first predicted step, and goal graphs
specifying the desired views. It learns subsequent graph and pose changes from
generated reference trajectories. The planner recursively extends candidate
paths and checks for predicted goal-graph matches. It selects the shortest
predicted goal-reaching path and follows it over a short execution horizon,
then repeats prediction and planning using newly observed SGs and probe poses.

The main contributions are:
\begin{itemize}
\item We propose \ours{}, an action- and goal-conditioned world model that jointly predicts anatomical SGs and probe poses for interpretable, anticipatory probe navigation. A receding-horizon planner selects the shortest predicted path reaching a goal graph and follows it over a short horizon before replanning using new observations.
\item We introduce a CT-derived training-data pipeline that generates aligned SG--pose sequences from anatomical label maps, reducing the need for tracked and anatomically annotated US training sequences.
\item We evaluate prediction and closed-loop navigation on held-out CT cases and demonstrate the feasibility of real-world robotic US navigation through phantom experiments.

\end{itemize}

\section{Related Work}
\label{sec:related}
\subsection{Autonomous Probe Navigation in Robotic Ultrasound}

Autonomous robotic US scanning combines control of probe contact and orientation with
navigation toward a target view~\cite{bi2024machine}. Force- and confidence-based
controllers address the contact and alignment requirements~\cite{jiang2020force,jiang2020normal}.
Building on these controls, navigation methods use anatomical geometry, image feedback,
or learned policies to determine how the probe should move.

\begin{figure*}[t]
  \centering

  \includegraphics[width=0.95\textwidth]{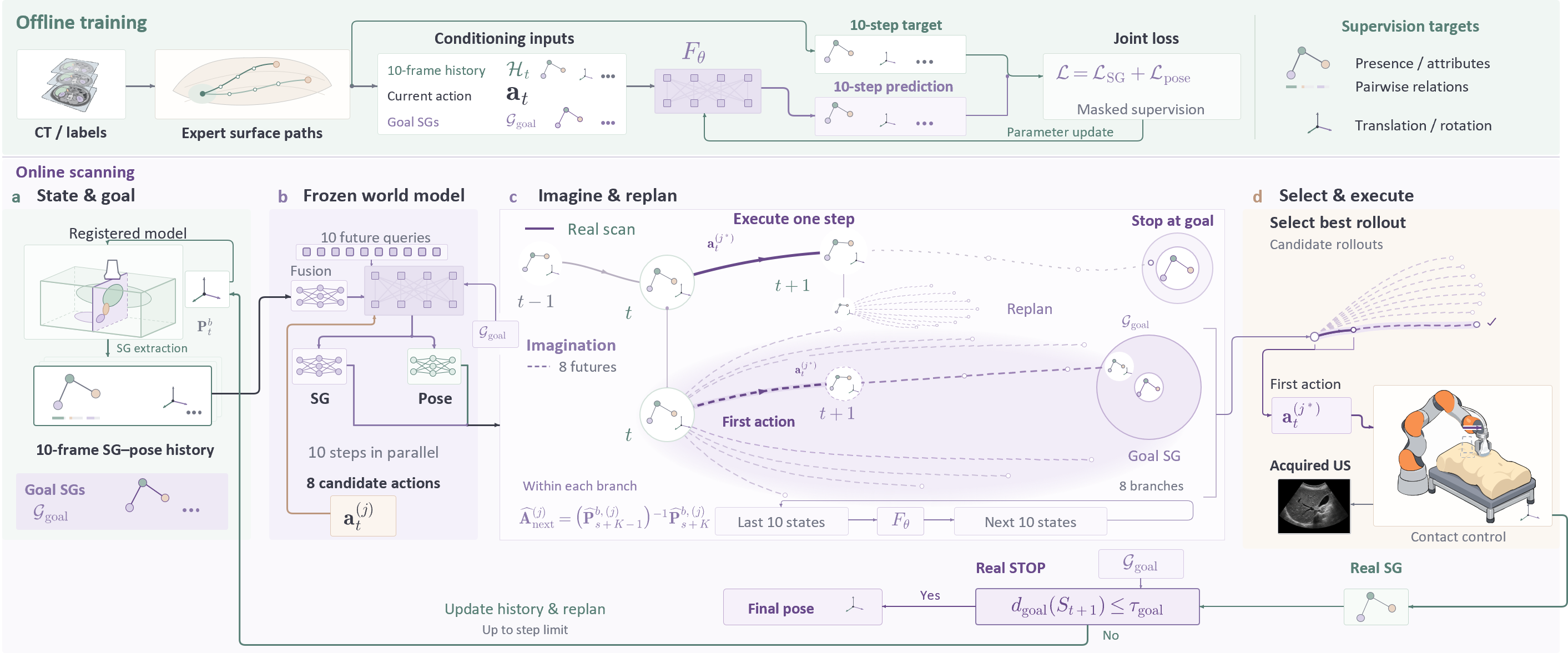}
  \caption{System overview of \ours. CT label maps supply SG--pose sequences for
           joint supervision. During navigation, the frozen model recursively predicts
           independent branches, each with an upper limit of 300 states.
           Branches stop at their first predicted goal match or an invalid state.
           Selection minimizes predicted probe travel among goal-matching branches;
           one action is executed before the observed SG is checked and history updated.}
  \label{fig:system}
\end{figure*}

Geometry-based methods transfer template scanning paths through rib-cartilage
US--CT registration~\cite{jiang2023skeleton,jiang2025cartilage} or use segmented
vessel geometry to maintain centering and perpendicular imaging during
scanning~\cite{jiang2022tubular}. View-quality methods guide target search through
Bayesian optimization with an expert-derived prior~\cite{raina2023sonographer}
or reward functions learned from ranked image pairs~\cite{jiang2024sonographer}.

Learned policies map observations to probe motions. Deep RL has been applied
to target-plane navigation~\cite{hase2020navigation,li2021autonomous},
including transfer from simulated vessel masks to real US
segmentations~\cite{bi2022vesnet}. SonoGym supports policy training and
evaluation in simulation~\cite{ao2025sonogym}, while UltraBot learns
probe adjustments through imitation of expert demonstrations~\cite{jiang2025ultrabot}.

While these approaches can guide probe motion based on the US observations, they generally do not explicitly predict how the anatomical view will evolve along candidate paths. A world model that predicts future anatomical views would enable the robot to compare candidate paths before execution and select those expected to reach the target view efficiently, providing an explicit, interpretable basis for navigation.

\subsection{World Models for Ultrasound}

A world model learns how actions change subsequent observations or representations of
the scene. The quantities it predicts depend on how they will be used: PlaNet plans
through learned latent states~\cite{hafner2019planet}, while MuZero predicts rewards
and values needed for planning~\cite{schrittwieser2020muzero}. Contrastive structured
world models learn how object-level states change under actions~\cite{kipf2020cswm}.
For US, this raises two related questions: what information about a future view should
be predicted, and how should that prediction guide probe movement?

One use of world modeling is to learn anatomical and motion information for a
probe-guidance network. EchoWorld predicts masked and motion-conditioned image features
during pretraining, then fine-tunes a guidance network that uses previous images and
motions~\cite{yue2025echoworld}. Fan et al.~\cite{fan2026acwm} train an
action-conditioned diffusion model to predict future US images. They then use rewards
from the frozen world model to fine-tune a goal-conditioned action predictor.
In these approaches, learning how images change with probe motion supports the training
of a network that produces probe guidance.

World-model predictions can also be used during inference, as illustrated by DreamReg
for 2D--3D US registration~\cite{kang2026dreamreg}. DreamReg summarizes previous
observations and poses in a latent estimate of the alignment, then simulates candidate
probe motions and their predicted observations to refine the registration transform.

Anticipating the consequences of probe motion does not necessarily require predicting every detail of a future US image. Sonographers can instead draw on anatomical knowledge to reason about which structures may appear and how their spatial arrangement may change. Given the noisy nature of US images, high-fidelity image prediction introduces the additional burden of modeling appearance variations that may be irrelevant to navigation. Scene graphs offer a practical alternative by explicitly representing anatomical structures, their geometry, and spatial relationships, allowing a world model to focus on navigation-relevant anatomical changes.

\subsection{Anatomical Scene Graphs and CT-Derived Training Data}

Scene graphs explicitly represent objects and their relationships, allowing image
content to be described through the objects present and how they are
arranged~\cite{johnson2015image}. In robotics, 3D dynamic scene graphs organize objects,
places, and their spatial relationships~\cite{rosinol2020dsg}; SayPlan uses these graphs
to identify objects and locations relevant to a task and plan robot
actions~\cite{rana2023sayplan}.
For US, Li et al.~\cite{li2025scenegraph} estimate anatomical structures and their
relations from neck images, then use the graphs for image explanation and scanning
guidance toward missing structures. Their work shows how the structures visible in a
US image can provide information for deciding where to scan next.

Predicting SGs along a probe path adds a temporal requirement to image-based SG
estimation. Li et al.~\cite{li2025scenegraph} annotate structures and relations in
individual neck US images and identify annotation effort and limited anatomical coverage
as constraints. World-model training additionally requires these labels across
successive scanning steps, paired with the probe's position and orientation.
Image-level graph annotations alone do not provide these SG--pose sequences.

CT segmentation resources offer a source of anatomical labels from which such
sequences can be constructed. TotalSegmentator provides CT data and segmentation models
covering multiple organs and other anatomical structures~\cite{wasserthal2023totalsegmentator}.
CADS combines and standardizes CT datasets to provide labels with broad anatomical
coverage~\cite{xu2025cads}. These volume labels allow structures and their relationships
to be extracted at successive imaging planes along a probe path.
We use CADS label maps to construct SGs at sampled probe poses and assemble them into
training sequences. The resulting SG--pose pairs let the world model learn how
structures in the imaging plane change as the probe moves. During inference, SGs can be extracted from acquired US images using specialized anatomical detection models.

\section{Methods}
\label{sec:method}

\subsection{Problem Formulation}

\ours{} predicts anatomical views and probe poses to compare candidate motions before
execution (Fig.~\ref{fig:system}). At time $t$, $\mathcal H_t$ contains the latest $H$
observed pairs $(S_i,\mathbf P_i^b)$, including the current state.
Here, $S_i$ denotes the observed SG.
The pose $\mathbf P_i^b\in SE(3)$ maps the
probe frame to the anatomy-specific body frame. Consecutive poses define the local motion
\begin{equation}
\mathbf A_t=(\mathbf P_t^b)^{-1}\mathbf P_{t+1}^b,
\label{eq:action_transform}
\end{equation}
so $\mathbf P_{t+1}^b=\mathbf P_t^b\mathbf A_t$. This gives action labels during training;
at inference, candidate actions are proposed before execution. Their six-dimensional
encoding $\mathbf a_t$ comprises local translation (mm) and a rotation vector (rad).

The goal is an unordered set of acceptable views,
$\mathcal G_{\mathrm{goal}}=\{S_m^{\mathrm{goal}}\}_{m=1}^{M}$, supplied for the requested
target. Goal poses are not model inputs; goal-set construction and availability are
specified in Sec.~\ref{sec:experimental_setup}. The world model predicts
\begin{equation}
\widehat{\mathcal T}_t
=F_\theta(\mathcal H_t,\mathbf a_t,\mathcal G_{\mathrm{goal}}),
\label{eq:world_model}
\end{equation}
where $\widehat{\mathcal T}_t$ contains $K$ predicted pairs
$(\widehat S_{t+k},\widehat{\mathbf P}^b_{t+k})$ for $k=1,\ldots,K$.
Each graph describes the anatomy expected at its paired pose.
The candidate action conditions the first transition. Later steps form a learned
goal-conditioned continuation.

\subsection{Ultrasound Scene-Graph World Model Architecture}
\label{sec:sg_state}
\label{sec:world_model}

\begin{figure}[t]
  \centering
  \includegraphics[width=0.8\columnwidth]
    {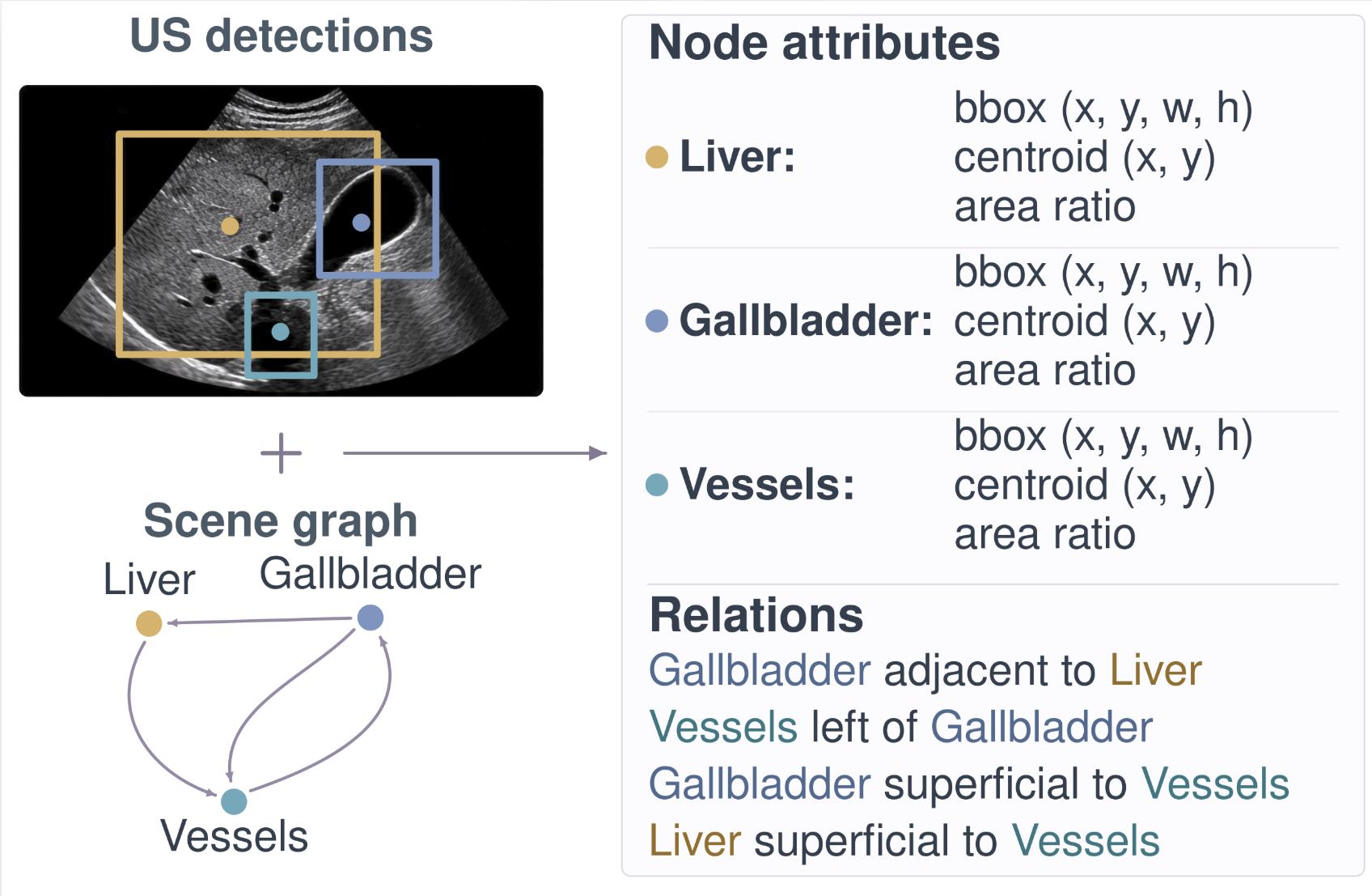}
  \caption{
    Ultrasound scene graph representation with per-node attributes
    and relations.
  }
  \label{fig:us_scene_graph}
\end{figure}

The world model jointly predicts future anatomical SGs and probe poses,
conditioned on the observed SG--pose history, a candidate action,
and a set of goal SGs. Each SG describes the anatomical structures
intersecting the imaging plane. Nodes have fixed anatomical identities
and eight numerical attributes: component count, normalized bounding-box
coordinates $(x,y,w,h)$, normalized centroid coordinates $(x,y)$,
and visible-area ratio. Spatial relationships between structures are
represented by three relation types: \emph{adjacent to},
\emph{left of}, and \emph{superficial to}. One example is shown in Fig.~\ref{fig:us_scene_graph}.

A shared SG encoder combines node features and pairwise features,
including spatial relations, into a $d$-dimensional embedding.
Probe poses are encoded from a 9-D representation comprising 3-D
translation and continuous 6-D rotation. A multilayer perceptron
fuses each pair of SG and pose embeddings into a state token.
The Transformer encoder takes $H$ history state tokens, one candidate
action token, and $M$ goal tokens obtained using the shared SG encoder.
Temporal embeddings encode the order of the history states, while
type embeddings distinguish the action and goal tokens. Goal tokens
receive no order-dependent positional encoding, allowing the goals
to be represented as an unordered set.

The Transformer decoder uses $K$ learned queries to predict the next
$K$ SG--pose pairs in parallel. Each decoded token is passed to an
SG head and a pose head. The SG head predicts node presence, numerical
attributes, pair features, and spatial relations under the fixed graph
schema. The pose head predicts relative 3-D translation and continuous
6-D rotation, which are converted into a rigid transformation
$\widehat{\Delta\mathbf P}_{t\rightarrow t+k}\in SE(3)$.
The future probe pose is reconstructed as $\widehat{\mathbf P}^{b}_{t+k}=\mathbf P_t^b \widehat{\Delta\mathbf P}_{t\rightarrow t+k}.$
All predicted poses within a chunk are expressed relative to the same
current pose $\mathbf P_t^b$. Accordingly, the supervision target
for step $k$ is $\Delta\mathbf P_{t\rightarrow t+k}=(\mathbf P_t^b)^{-1}\mathbf P_{t+k}^b.$

\subsection{CT-Derived Supervision and Training}
\label{sec:data_preparation}
\label{sec:training_objective}

We train on SG--pose sequences generated from CADS label maps~\cite{xu2025cads};
trajectory construction and dataset settings are given in Sec.~\ref{sec:experimental_setup}.
Predicting anatomical structure rather than image texture lets CT labels supply this
supervision without paired annotated US scans.
At each waypoint, the imaging plane intersects the label map to obtain the anatomical
structures and their spatial relationships. The same procedure generates history,
future, and goal SGs. The reference
trajectories are geometrically generated, not recorded sonographer demonstrations.

Each predicted SG--pose pair is supervised by the corresponding future waypoint in
the CT-derived trajectory. The training objective combines graph and pose losses:
\begin{equation}
\begin{aligned}
\mathcal L &= \mathcal L_{\mathrm{SG}}+\mathcal L_{\mathrm{pose}},\\
\mathcal L_{\mathrm{SG}} &= \mathcal L_{\mathrm{pres}}+\mathcal L_{\mathrm{attr}}
  +\mathcal L_{\mathrm{pair}}+\mathcal L_{\mathrm{rel}},\\
\mathcal L_{\mathrm{pose}} &= \mathcal L_{\mathrm{trans}}+\mathcal L_{\mathrm{rot}}.
\end{aligned}
\label{eq:loss}
\end{equation}
$\mathcal L_{\mathrm{pres}}$ applies binary cross-entropy to each anatomical node's
presence label. $\mathcal L_{\mathrm{rel}}$ applies binary cross-entropy independently
to each relation label, allowing multiple relations per node pair.
$\mathcal L_{\mathrm{attr}}$ and $\mathcal L_{\mathrm{pair}}$ use SmoothL1 for numerical
node attributes and pair features, respectively; component count is regressed.
Pose supervision uses the relative transforms $\Delta\mathbf P_{t\rightarrow t+k}$.
Let $\mathbf R_k$ and $\widehat{\mathbf R}_k$ denote their reference and predicted
rotation matrices, respectively. The per-step rotation loss is $\ell_{\mathrm{rot},k}
=\arccos\!\left(\frac{\operatorname{tr}(\widehat{\mathbf R}_k^\top\mathbf R_k)-1}{2}\right).$

$\mathcal L_{\mathrm{rot}}$ aggregates these angles over valid future steps, while
$\mathcal L_{\mathrm{trans}}$ penalizes relative translation error.

\subsection{Planning by Recursive Imagination}
\label{sec:planner}

The planner uses the world model to compare candidate probe trajectories before
execution. At each planning iteration, it generates candidate motions, recursively
predicts their anatomical outcomes, and selects the shortest predicted path that
reaches a goal view. The robot follows the selected trajectory over a short execution
horizon before replanning from new observations.

At time $t$, we sample $C$ candidate probe motions by randomly perturbing the current
probe velocity. Each candidate initializes an independent imagined branch, conditioned
on the same observed history $\mathcal H_t$ and goal set. The world model extends each
branch in $K$-step chunks. An observed or predicted SG matches the goal set if its distance to
at least one goal graph falls below a threshold calibrated on training
cases. The graph distance compares node presence, numerical attributes,
pair features, and spatial relations.

For branch $j$, a chunk starting at index $s$ predicts SGs and body-frame probe poses
at $s+1,\ldots,s+K$, where $s=t$ for the first chunk. To continue the rollout, we derive
the next conditioning motion from the final two predicted poses:
\begin{equation}
\widehat{\mathbf A}_{\mathrm{next}}^{(j)}
=
(\widehat{\mathbf P}_{s+K-1}^{b,(j)})^{-1}
\widehat{\mathbf P}_{s+K}^{b,(j)}.
\label{eq:recursive_action}
\end{equation}
Its 6-D encoding serves as the first-action input for the next chunk, starting at
$s+K$. The branch history is updated using the latest $H$ SG--pose pairs, including
predicted pairs, while the goal set remains fixed. This process extends the imagined
trajectory without acquiring new observations. Predicted states are checked in temporal
order, and each branch terminates at its first goal match, immediately before an invalid
state, or upon reaching the imagination limit $n_{\max}$.

Let $\mathcal B_{\mathrm{goal}}$ denote the branches that reach a predicted
goal within the imagination limit, and let $n_j$ denote the first
goal-matching step of branch $j$. The cumulative translational distance
along this branch to the predicted goal is
\begin{equation}
D_j
=
\sum_{k=1}^{n_j}
\left\|
\widehat{\mathbf p}^{b,(j)}_{t+k}
-
\widehat{\mathbf p}^{b,(j)}_{t+k-1}
\right\|_2,
\label{eq:predicted_path_length}
\end{equation}
where $\widehat{\mathbf p}^{b,(j)}_{t+k}$ is the translational component
of the predicted probe pose, and
$\widehat{\mathbf p}^{b,(j)}_t=\mathbf p_t^b$ is the current observed
position. The planner selects the goal-reaching branch with the
shortest predicted travel distance:
\begin{equation}
j^*
=
\arg\min_{j\in\mathcal B_{\mathrm{goal}}} D_j.
\label{eq:shortest_branch_selection}
\end{equation}
Thus, selection minimizes predicted translational travel to a goal, measured in
millimeters. Probe orientation is included in the predicted poses but does not
contribute to the path-length criterion. If no branch reaches a predicted goal, the
planner returns \textsc{No\_Goal\_Reached} without selecting a trajectory.

The robot follows the selected trajectory under the surface constraint over a short
execution horizon. Newly acquired SGs and measured probe poses then update the observed
history for the next planning iteration. Navigation succeeds only when an observed SG
satisfies $d_{\mathrm{goal}}(S)\leq\tau_{\mathrm{goal}}$; a predicted goal match alone
does not establish task completion. The prediction, selection, and execution cycle
repeats until observed success, no eligible branch, or the execution limit is reached.

\section{Experimental Evaluation}
\label{sec:experiments}

\subsection{Data and Implementation Details}
\label{sec:experimental_setup}

\subsubsection{Dataset and Reference Goals}

We use CT volumes and corresponding anatomical label maps from CADS~\cite{xu2025cads}.
From 24 patient cases, we generate 1,839 probe trajectories containing 63,164 waypoints.
For each patient CT and target organ, we predefine probe positions and orientations
that yield acceptable views of that organ, termed \emph{target-view poses}.
We extract a reference SG from the CT label map at each of these poses.
By default, each target organ has $M=3$ reference SGs representing
alternative acceptable views.
The complete set of reference SGs is supplied as $\mathcal G_{\mathrm{goal}}$ at
inference, including for held-out patients.

Training, future-state prediction, and CT-based navigation use annotation-derived SGs.
The robot--phantom protocol generates SG observations from a registered phantom label
map and measured probe poses. These settings test prediction and planning separately
from B-mode-to-SG estimation. The graph schema uses at most seven anatomical nodes,
21 node pairs, and 51 relation triplets. Pair geometry comprises horizontal displacement,
relative depth, normalized distance, and an adjacency score.

\subsubsection{Geometric Reference Trajectory Generation}

For each patient, we first select and manually validate a set of clinically meaningful target views for each target organ. We then sample diverse initial probe poses across the accessible surface of the lower abdomen and manually exclude infeasible or highly redundant configurations. Each retained start pose is paired with every verified target view for the corresponding organ, and a smooth reference trajectory constrained to the body surface is generated. Along each trajectory, the probe progressively adjusts its position and orientation toward the target view. All trajectories undergo quality control for contact validity, geometric feasibility, motion continuity, and terminal consistency, with manual review and correction when necessary. At each sampled waypoint, the corresponding scene graph is constructed based on CT label maps, yielding synchronized SG--pose pairs (Sec.~\ref{sec:data_preparation}). The transformation between consecutive poses is represented as a relative action in the probe-local coordinate frame (Eq.~\eqref{eq:action_transform}). Each expert trajectory therefore provides a temporally aligned sequence of SG observations, probe poses, and local actions for subsequent world model training.

\begin{figure*}[t]
  \centering
  \includegraphics[width=0.85\textwidth]{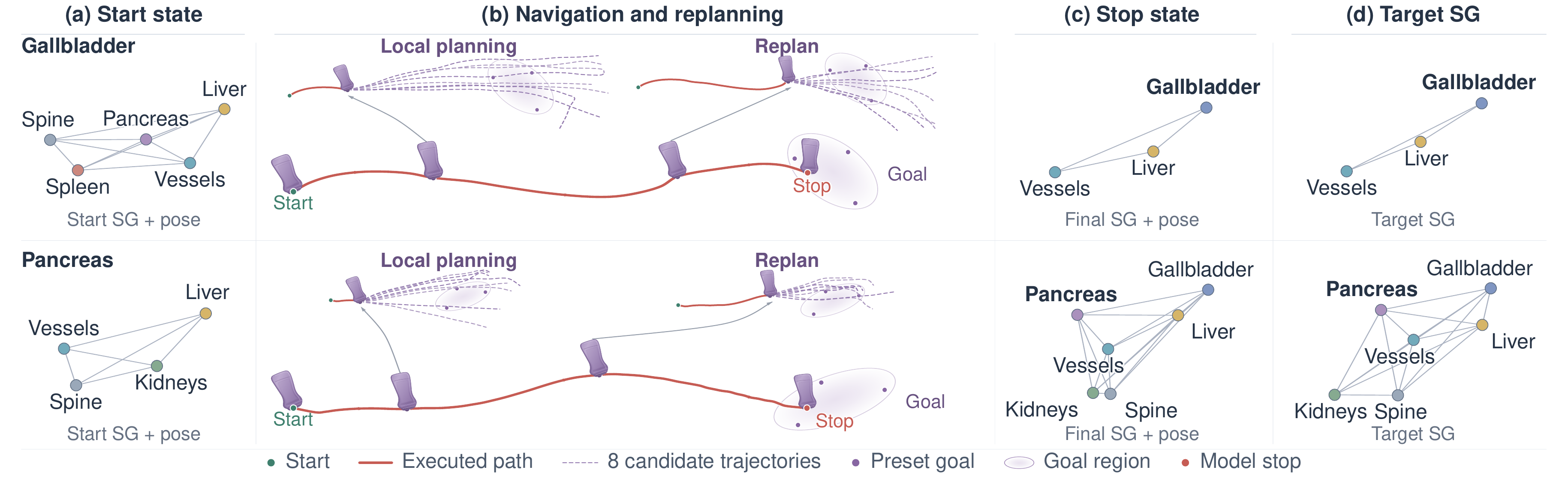}
  \caption{
Closed-loop navigation examples for the gallbladder (top) and pancreas (bottom). 
(a) Initial SG and probe pose. 
(b) \ours evaluates eight candidate future trajectories, executes the first action of the selected trajectory, and replans from the updated observation. Purple dashed curves show predicted candidates and red curves show the executed path. 
(c) Final SG and probe pose at the model-selected stopping state. 
(d) Goal SG used for navigation conditioning.
}
  \label{fig:navigation_cases}
\end{figure*}

\subsubsection{Temporal Samples and Case Split}

Each sample contains $H=10$ history SG--pose pairs, the current trajectory action,
the target-specific goal set, and $K=10$ future SG--pose targets. 

The patient-disjoint split comprises 20 training cases (53,344 samples) and four test cases (7,981 samples). Normalization statistics and the goal threshold use training cases only.

\subsubsection{Model and Planner Configuration}

The Transformer uses embedding dimension $d=384$, four encoder layers, four decoder
layers, eight attention heads, a feed-forward dimension of 1,536, and dropout of 0.1.

Action vectors and pose translations are normalized using training-set statistics
before network encoding. This preprocessing is omitted from the model notation;
execution uses physical units.

At each planning iteration, the planner samples $C=8$ candidate motions by randomly perturbing the current probe velocity. Each branch can extend to at most $n_{\max}=300$ predicted states
(30 chunks of $K=10$). Goal matching or an invalid state terminates a branch earlier.
Selection uses the predicted path length up to each eligible branch's first goal match;
each planning iteration executes one physical step.
Because $H=K$ in this configuration, each recursive call uses the previous chunk's
predicted SG--pose pairs as its history.

\subsubsection{Robotic Ultrasound Setup}
\label{sec:hardware_setup}

The robotic US platform comprises an LBR iiwa 7 R800 robot (KUKA, Germany) and a
bk3000 US system (BK Medical, Herlev, Denmark) equipped with a 5C1e
curved-array transducer. The probe is mounted on the robot end effector, and the
B-mode imaging depth is set to 16\,cm.
Robot experiments use a US-22 CT/US abdominal phantom (Kyoto Kagaku, Japan).

\begin{table*}[t]
\caption{Scene-graph prediction accuracy on held-out cases with trajectory-action
inputs.
         }
\label{tab:prediction}
\centering
\footnotesize
\setlength{\tabcolsep}{4pt}
\begin{tabular}{lcccccccc}
\toprule
 & & & \multicolumn{3}{c}{\textbf{Gallbladder}}
     & \multicolumn{3}{c}{\textbf{Pancreas}} \\
\cmidrule(lr){4-6}\cmidrule(lr){7-9}
Prediction stage
  & \shortstack{Local step\\$k$}
  & \shortstack{Total steps\\$n$}
  & \shortstack{Node visibility\\F1(\%) $\uparrow$}
  & \shortstack{Spatial relation\\F1(\%) $\uparrow$}
  & \shortstack{BBox mIoU\\(\%) $\uparrow$}
  & \shortstack{Node visibility\\F1(\%) $\uparrow$}
  & \shortstack{Spatial relation\\F1(\%) $\uparrow$}
  & \shortstack{BBox mIoU\\(\%) $\uparrow$} \\
\midrule
Initial   &  1 &  1 & $88.74 \pm 0.21$ & $94.45 \pm 0.08$ & $39.92 \pm 0.32$ & $93.13 \pm 0.19$ & $94.89 \pm 0.10$ & $42.44 \pm 0.37$ \\
          &  3 &  3 & $87.69 \pm 0.23$ & $94.35 \pm 0.08$ & $38.95 \pm 0.34$ & $92.33 \pm 0.22$ & $94.77 \pm 0.11$ & $41.13 \pm 0.40$ \\
          &  5 &  5 & $86.95 \pm 0.24$ & $94.20 \pm 0.09$ & $38.34 \pm 0.34$ & $91.42 \pm 0.26$ & $94.67 \pm 0.11$ & $39.79 \pm 0.42$ \\
          & 10 & 10 & $84.84 \pm 0.30$ & $93.80 \pm 0.13$ & $36.09 \pm 0.42$ & $89.97 \pm 0.33$ & $94.69 \pm 0.11$ & $38.87 \pm 0.49$ \\
\midrule
Recursive &  1 & 11 & $84.21 \pm 0.31$ & $94.14 \pm 0.09$ & $34.39 \pm 0.37$ & $89.21 \pm 0.33$ & $94.26 \pm 0.15$ & $37.00 \pm 0.50$ \\
          &  3 & 13 & $84.10 \pm 0.34$ & $94.16 \pm 0.10$ & $33.57 \pm 0.43$ & $89.07 \pm 0.38$ & $94.19 \pm 0.15$ & $35.60 \pm 0.56$ \\
          &  5 & 15 & $83.94 \pm 0.37$ & $94.12 \pm 0.12$ & $32.92 \pm 0.46$ & $88.57 \pm 0.42$ & $94.28 \pm 0.16$ & $34.60 \pm 0.60$ \\
          & 10 & 20 & $82.70 \pm 0.51$ & $93.59 \pm 0.18$ & $31.58 \pm 0.57$ & $87.50 \pm 0.54$ & $94.19 \pm 0.19$ & $34.09 \pm 0.70$ \\

\bottomrule
\end{tabular}
\end{table*}

\subsection{Evaluation Questions and Common Protocol}

The evaluation addresses three questions: (1) How accurately do predicted SGs represent the anatomy at their corresponding predicted probe poses, including over recursively extended rollouts? (2) How reliably does the planner reach an acceptable target view from different initial distances? (3) Can the system guide a physical robot to the target view in phantom experiments? We evaluate navigation toward two target organs, gallbladder and pancreas. For each target $q$, $\mathcal G_{\mathrm{goal}}^{(q)}$ denotes the set of all acceptable goal graphs.

Navigation trials use a frozen world model, fixed planner settings, and a fixed execution budget. Initial poses are divided into near ($d_0\leq d_{\mathrm{split}}$) and far ($d_0>d_{\mathrm{split}}$) groups according to their reference surface-path distance $d_0$ to the nearest hidden goal pose. The threshold $d_{\mathrm{split}}$ is set before evaluation on held-out cases. The goal poses are used only for trial sampling and evaluation and are not available to the planner. Results are reported for each target and distance group, together with an overall summary.

\begin{table*}[t]
\caption{Closed-loop target-view acquisition on held-out CT cases with execution
horizons of 1, 5, and 10 steps.}
\label{tab:navigation}
\centering
\footnotesize
\setlength{\tabcolsep}{3pt}

\resizebox{\textwidth}{!}{
\begin{tabular}{lcccccccccccc}
\toprule
&
\multicolumn{4}{c}{Execute 1} &
\multicolumn{4}{c}{Execute 5} &
\multicolumn{4}{c}{Execute 10} \\
\cmidrule(lr){2-5}
\cmidrule(lr){6-9}
\cmidrule(lr){10-13}

Initial distance
  & \shortstack{Successes / trials\\$n/N$}
  & \shortstack{Success rate\\(\%) $\uparrow$}
  & \shortstack{Actions to goal\\(steps) $\downarrow$}
  & \shortstack{Final position error\\(mm) $\downarrow$}
  & \shortstack{Successes / trials\\$n/N$}
  & \shortstack{Success rate\\(\%) $\uparrow$}
  & \shortstack{Actions to goal\\(steps) $\downarrow$}
  & \shortstack{Final position error\\(mm) $\downarrow$}
  & \shortstack{Successes / trials\\$n/N$}
  & \shortstack{Success rate\\(\%) $\uparrow$}
  & \shortstack{Actions to goal\\(steps) $\downarrow$}
  & \shortstack{Final position error\\(mm) $\downarrow$} \\
\midrule

\multicolumn{13}{l}{\textbf{Gallbladder}} \\

Near
& 52/64  & 81.25 & $19.37 \pm 11.76$ & $12.43 \pm 11.86$
& 22/64  & 34.38 & $19.64 \pm 15.84$ & $15.32 \pm 13.29$
& 25/64  & 39.06 & $27.64 \pm 28.30$ & $11.69 \pm 9.60$ \\

Far
& 41/56  & 73.21 & $37.51 \pm 15.67$ & $16.25 \pm 11.77$
& 27/56  & 48.21 & $38.59 \pm 16.32$ & $12.14 \pm 9.08$
& 21/56  & 37.50 & $42.76 \pm 25.40$ & $13.38 \pm 11.04$ \\

Overall
& 93/120 & 77.50 & $27.37 \pm 16.29$ & $14.11 \pm 11.91$
& 49/120 & 40.83 & $30.08 \pm 18.57$ & $13.56 \pm 11.16$
& 46/120 & 38.33 & $34.54 \pm 27.78$ & $12.46 \pm 10.20$ \\

\midrule

\multicolumn{13}{l}{\textbf{Pancreas}} \\

Near
& 52/65  & 80.00 & $18.33 \pm 10.57$ & $15.15 \pm 8.11$
& 19/65  & 29.23 & $13.95 \pm 22.02$ & $15.80 \pm 9.58$
& 21/65  & 32.31 & $17.57 \pm 21.42$ & $12.38 \pm 7.97$ \\

Far
& 38/55  & 69.09 & $33.18 \pm 13.56$ & $13.10 \pm 7.08$
& 9/55   & 16.36 & $46.89 \pm 34.60$ & $19.77 \pm 3.52$
& 6/55   & 10.91 & $58.00 \pm 35.01$ & $16.80 \pm 2.55$ \\

Overall
& 90/120 & 75.00 & $24.60 \pm 13.96$ & $14.28 \pm 7.72$
& 28/120 & 23.33 & $24.54 \pm 30.39$ & $17.08 \pm 8.27$
& 27/120 & 22.50 & $26.56 \pm 29.70$ & $13.36 \pm 7.33$ \\

\bottomrule
\end{tabular}
}
\end{table*}

\subsection{Scene-Graph Prediction Accuracy}
\label{sec:prediction_evaluation}

We evaluate the world model's ability to predict anatomical SGs over an
initial prediction chunk and its first recursive extension on four held-out
CT cases. Rollouts are conditioned on either a gallbladder or pancreas
goal set, with both groups evaluated over the full anatomical graph.
Each rollout begins with a ten-frame reference history and the corresponding
trajectory action. The recursive extension uses the model's predicted
SG--pose history and the motion derived from
Eq.~\eqref{eq:recursive_action}, without additional reference observations.
We evaluate local steps $k\in\{1,3,5,10\}$ in each chunk, covering cumulative
prediction steps up to $n=20$.

For each predicted SG--pose pair, we extract a reference SG from the
held-out CT label map at the predicted pose. This comparison assesses
whether the imagined anatomy agrees with the predicted probe location.
We report three complementary metrics: \emph{node visibility F1} measures
which anatomical structures are present, \emph{spatial relation F1}
measures their pairwise relationships, and \emph{BBox mIoU} measures
the overlap between predicted and reference anatomical bounding boxes.
Table~\ref{tab:prediction} reports the mean and standard deviation
of each metric at each evaluated step.

Spatial relation prediction remains stable throughout the evaluated
horizon. Between steps 1 and 20, relation F1 decreases only from 94.45\%
to 93.59\% for gallbladder-conditioned rollouts and from 94.89\% to
94.19\% for pancreas-conditioned rollouts. In contrast, node visibility
F1 decreases from 88.74\% to 82.70\% and from 93.13\% to 87.50\%,
respectively. Geometric localization also deteriorates: BBox mIoU
decreases from 39.92\% to 31.58\% for the gallbladder group and from
42.44\% to 34.09\% for the pancreas group. Both visibility and bounding-box
overlap continue to decline during recursive extension.

These results suggest that the model retains the broad spatial
relationships between anatomical structures over extended imagined
rollouts, while predicting their visibility and precise geometry
becomes more difficult. The gap between high relation F1 and limited
bounding-box overlap identifies geometric precision as a remaining
challenge for long-horizon prediction. Whether these predictions
support target-view acquisition is assessed separately through
closed-loop navigation experiments.

\subsection{Closed-Loop Navigation Performance}

We evaluate closed-loop target-view acquisition in held-out CT anatomical
environments for the gallbladder and pancreas. To examine the effect of
replanning frequency, we compare execution horizons of 1, 5, and 10 steps,
denoted \emph{Execute 1}, \emph{Execute 5}, and \emph{Execute 10}.
At each planning iteration, the robot follows the selected predicted
trajectory for the specified horizon before updating its SG--pose
history from new observations and replanning. A trial succeeds when an
observed SG satisfies the goal criterion within the execution budget.

Table~\ref{tab:navigation} reports results for 120 trials per target
under each execution setting, grouped by near and far initial distances.
Success rate is the proportion of all attempted trials that reach an
acceptable target view. Actions to goal count the executed probe-motion
steps until the first successful observation. Final position error measures the minimum Euclidean distance from the terminal probe position to the region defined by the three predefined target-view positions. Action counts and
position errors are reported as mean $\pm$ standard deviation over
successful trials only. Overall action counts and position errors are calculated across all successful trials from both distance groups. Navigation success is determined by whether the observed SG matches an acceptable goal graph; position error separately measures the distance to the nearest reference target-view position.

Replanning after every step achieves the highest success rates for
both targets and in both distance groups. With \emph{Execute 1},
the planner succeeds in 93/120 gallbladder trials (77.50\%) and
90/120 pancreas trials (75.00\%). Increasing the execution horizon
to five steps reduces overall success to 40.83\% and 23.33\%,
respectively. With ten-step execution, success decreases further
to 38.33\% and 22.50\%. These results demonstrate the importance
of frequent observation updates for navigation with the current
world model. The degradation with longer execution horizons is
consistent with the reduced prediction accuracy observed over
extended rollouts.

Under \emph{Execute 1}, distant starts are associated with lower
success rates and longer successful trajectories. Gallbladder
success decreases from 81.25\% for near starts to 73.21\% for far
starts, while the mean action count increases from 19.37 to 37.51.
For the pancreas, success decreases from 80.00\% to 69.09\%,
and the mean action count increases from 18.33 to 33.18.
Longer execution horizons particularly affect pancreas navigation
from far starts, where success falls to 16.36\% with
\emph{Execute 5} and 10.91\% with \emph{Execute 10}.

For successful trials under \emph{Execute 1}, the mean final
position error is 14.11\,mm for the gallbladder and 14.28\,mm
for the pancreas. Thus, reaching an acceptable anatomical view
does not require exact recovery of a predefined reference position.
Although some longer-horizon settings yield lower mean position
errors, these values describe only the successful subset and
should be interpreted alongside their substantially lower
success rates.

Overall, the results support the use of predicted anatomical
trajectories for target-view acquisition, while showing that
reliable performance depends strongly on frequent replanning
from observed anatomy.

\subsection{Robot--Phantom Navigation Performance}

We further evaluate \ours{} on the US-22 phantom using the robotic US platform
described in Sec.~\ref{sec:hardware_setup}. A compliant position--force controller
maintains acoustic contact and workspace limits. The phantom, robot-base, and probe frames are
registered before each session. A registered phantom label map and the measured probe
pose generate the current SG, matching the similar SG  used for training.
The acquired B-mode stream is recorded for target-view and contact assessment but is not
an input to the planner. This experiment therefore tests physical execution and repeated
replanning, not B-mode-to-SG perception. Trials start from free poses across phantom surface.

\begin{figure}[b]
  \centering
  \includegraphics[width=1\columnwidth]
    {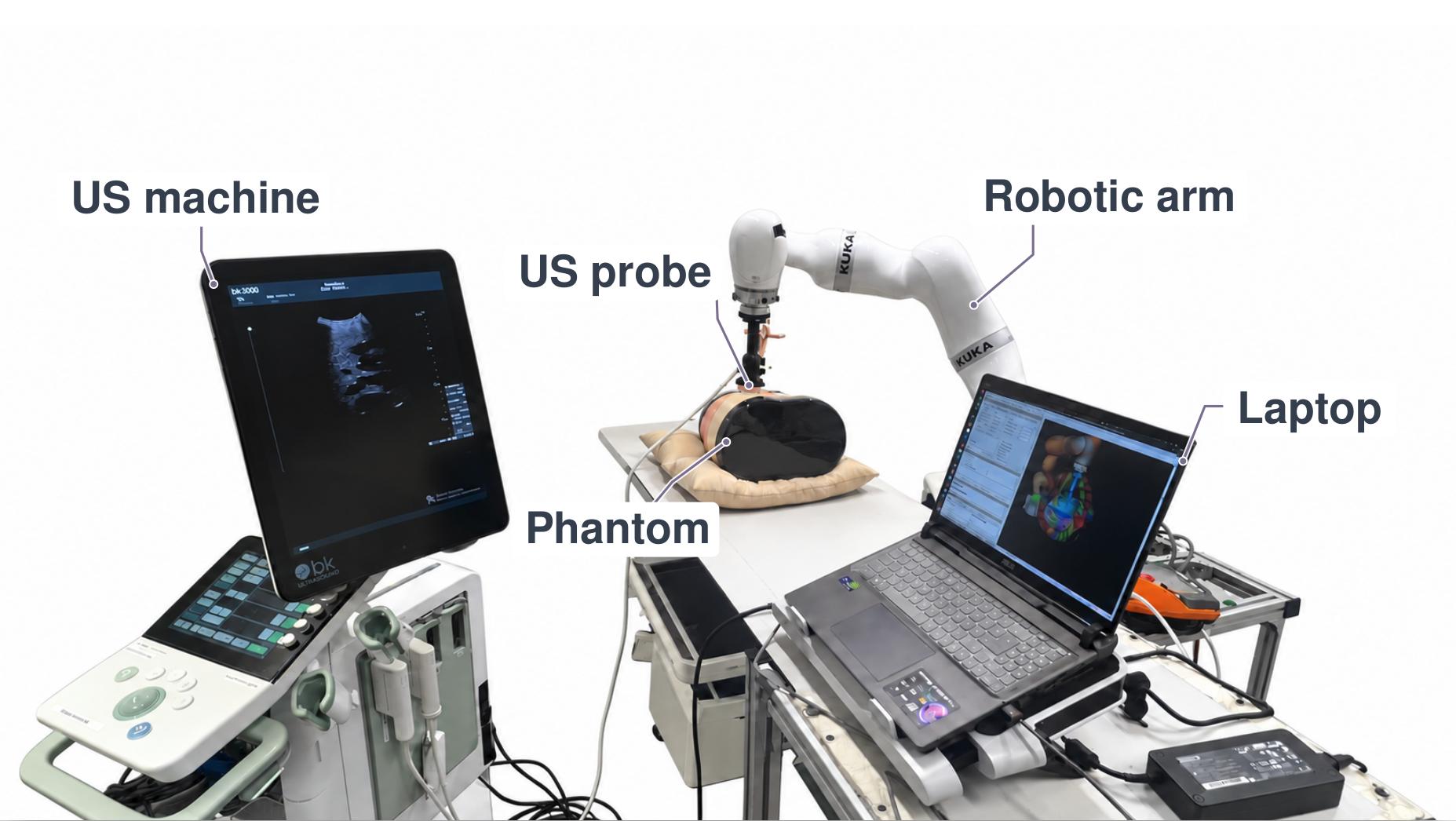}
  \caption{
    Experimental setup for scene graph-guided robotic ultrasound
    navigation on an abdominal phantom.
  }
  \label{fig:experimental_setup}
\end{figure}

The planner reached the target view in 8/10 gallbladder trials (80.0\%) and
6/9 pancreas trials (66.7\%), yielding an overall success rate of 73.7\% (14/19).
Among successful trials, actions to goal were $22.0 \pm 12.81$ steps for the
gallbladder and $23.5 \pm 12.44$ steps for the pancreas, with
$22.64 \pm 12.18$ steps overall (mean $\pm$ standard deviation).
Across all trials, final position errors were $37.70 \pm 19.08$\,mm for the
gallbladder and $23.07 \pm 10.45$\,mm for the pancreas (mean $\pm$ standard
deviation), with an overall mean of 30.77\,mm.

\FloatBarrier

\section{Conclusion and Limitations}
\label{sec:conclusion}

\ours{} learns anatomical transitions from CT-derived SG--pose sequences and
selects the shortest predicted path among branches that match a goal view.
On four held-out cases, spatial relation F1 remained above 93\% through 20 predicted
steps, while bounding-box overlap declined with the prediction horizon.
Closed-loop acquisition succeeded in 77.50\% of gallbladder trials and 75.00\% of pancreas trials.
Robot--phantom trials reached the target view in 14/19 cases (73.7\%), showing that the
predicted trajectories can be executed on a physical system.
These results support anatomical prediction and navigation with annotation-derived
SG observations and case-specific goal graphs.

Stable relation prediction coexisted with limited geometric precision.
Far-start trials had lower success rates and required more actions among successful trials.
Prediction accuracy beyond 20 steps remains uncharacterized despite a maximum planning
horizon of 300 states. Predicted goal matches guide selection, while observed SG
feedback verifies actual completion.

Evaluation covers four test anatomies and assesses the complete system without
isolating individual design contributions. Extending the approach to B-mode input
requires a schema-compatible SG estimator and evaluation of perception errors during navigation.
This would test whether CT-trained structural predictions remain useful with image-derived
observations. Cross-subject goal representations would further reduce reliance on
case-specific reference graphs.

\bibliographystyle{IEEEtran}
\balance
\bibliography{IEEEabrv,references}

\end{document}